\documentclass[10pt,twocolumn,letterpaper]{article}

\makeatletter
\def\@fnsymbol#1{\ensuremath{\ifcase#1\or \dagger\or \ddagger\or
   \mathsection\or \mathparagraph\or \|\or **\or \dagger\dagger
   \or \ddagger\ddagger \else\@ctrerr\fi}}
\makeatother

\usepackage{iccv}
\usepackage{times}
\usepackage{epsfig}
\usepackage{graphicx}
\usepackage{array,multirow}
\usepackage{amsmath}
\usepackage{amssymb}
\usepackage[title]{appendix}
\usepackage[ruled,vlined]{algorithm2e}

\usepackage[breaklinks=true,bookmarks=false]{hyperref}

\iccvfinalcopy 

\def\iccvPaperID{****} 

\ificcvfinal\fi

\begin{document}

\title{ReDAL: Region-based and Diversity-aware Active Learning for Point Cloud Semantic Segmentation}

\author{
Tsung-Han Wu$^1$ \qquad Yueh-Cheng Liu$^1$\thanks{Co-second authors contribute equally.} \qquad Yu-Kai Huang$^1$\footnotemark[1] \\
{Hsin-Ying Lee$^1$ \qquad Hung-Ting Su$^1$ \qquad Ping-Chia Huang$^1$ \qquad Winston H. Hsu$^{1,2}$} 
\\
\\
$^1$National Taiwan University \qquad $^2$Mobile Drive Technology
}

\maketitle
\ificcvfinal\thispagestyle{empty}\fi

\begin{abstract}
Despite the success of deep learning on supervised point cloud semantic segmentation, obtaining large-scale point-by-point manual annotations is still a significant challenge. To reduce the huge annotation burden, we propose a Region-based and Diversity-aware Active Learning (ReDAL), a general framework for many deep learning approaches, aiming to automatically select only informative and diverse sub-scene regions for label acquisition. Observing that only a small portion of annotated regions are sufficient for 3D scene understanding with deep learning, we use softmax entropy, color discontinuity, and structural complexity to measure the information of sub-scene regions. A diversity-aware selection algorithm is also developed to avoid redundant annotations resulting from selecting informative but similar regions in a querying batch. Extensive experiments show that our method highly outperforms previous active learning strategies, and we achieve the performance of 90\% fully supervised learning, while less than 15\% and 5\% annotations are required on S3DIS and SemanticKITTI datasets, respectively. Our code is publicly available at \href{https://github.com/tsunghan-wu/ReDAL}{\color{blue}{https://github.com/tsunghan-wu/ReDAL}}.
\end{abstract}

\section{Introduction}

Point cloud semantic segmentation is crucial for various emerging applications such as indoor robotics and autonomous driving. Many supervised approaches \cite{qi2017pointnet, qi2017pointnet++, wang2019graph, thomas2019kpconv, choy2019minkowski, tang2020spvconv} along with several large-scale datasets \cite{armeni2016s3dis, dai2017scannet, hackel2017semantic3d, behley2019semantickitti} are recently provided and have made huge progress.

\begin{figure}
    \centering
    \includegraphics[width=0.99\linewidth]{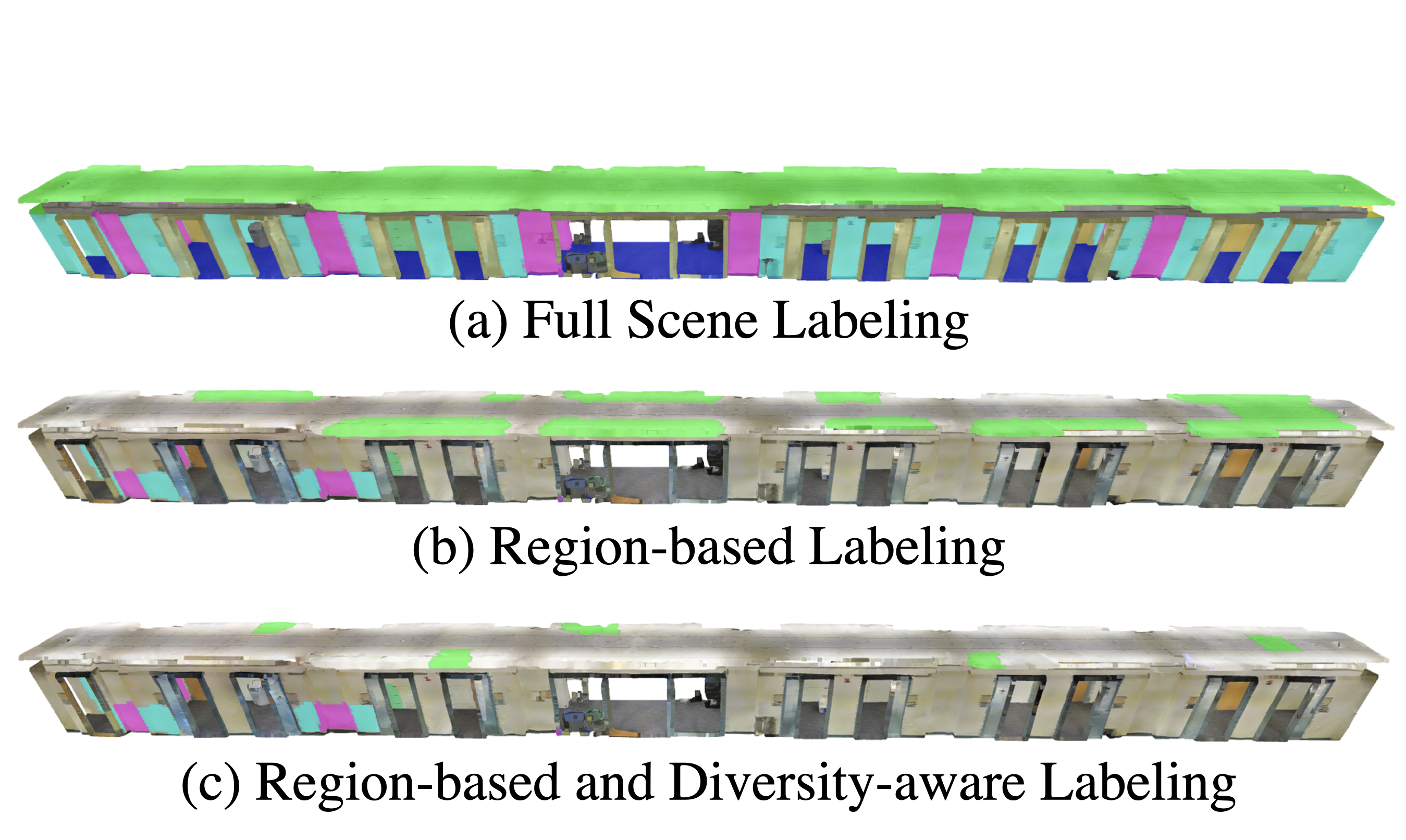}
    \caption{\textbf{Human labeling efforts (colored areas) of different learning strategies.} (a) In supervised training or traditional deep active learning, all points in a single point cloud are required to be labeled, which is labor-intensive. (b) Since few regions contribute to the model improvement, our region-based active learning strategy selects only a small portion of informative regions for label acquisition. Compared with case (a), our approach greatly reduces the cost of semantic labeling of walls and floors. (c) Moreover, considering the redundant labeling where repeating visually similar regions in the same querying batch, we develop a diversity-aware selection algorithm to further reduce redundant labeling (e.g., ceiling colored in green in (b) and (c)) effort by penalizing visually similar  regions.
    }
    \label{fig:fig1}
\end{figure}

Although recent deep learning methods have achieved great success with the aid of massive datasets, obtaining a large-scale point-by-point labeled dataset is still costly and challenging. Specifically, the statistics show that there would be more than 100,000 points in a room-sized point cloud scene \cite{armeni2016s3dis, behley2019semantickitti}. Furthermore, the annotation process of 3D point-wise data is much more complicated than that of 2D data. Unlike simply selecting closed polygons to form a semantic annotation in a 2D image \cite{russell2008labelme}, in 3D point-by-point labeling, annotators are asked to perform multiple 2D annotations from different viewpoints during the annotation process \cite{hackel2017semantic3d} or to label on 3D space with brushes through multiple zooming in and out and switching the brush size \cite{behley2019semantickitti}. Therefore, such numerous points and the complicated annotation process significantly increase the time and cost of manual point-by-point labeling.

\begin{figure*}[t!]
    \centering
    \includegraphics[width=\textwidth]{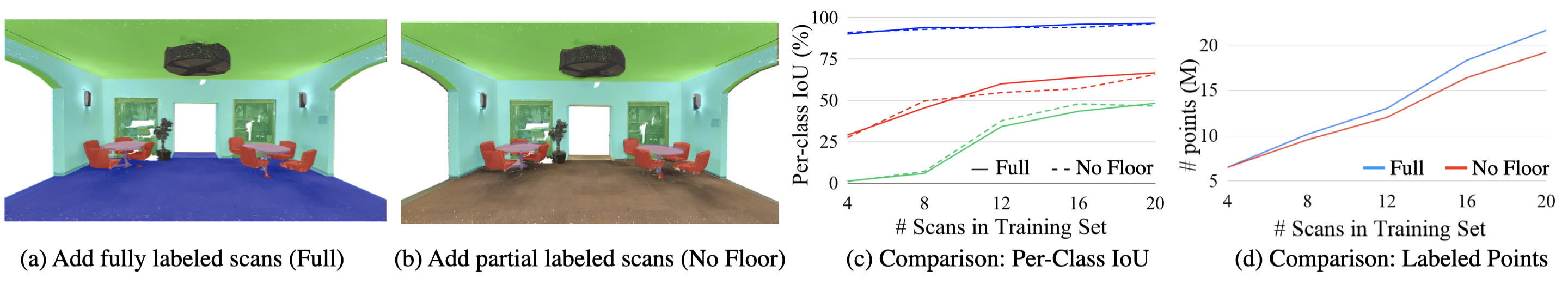}
    \caption{
    \textbf{Not all annotated regions contribute to the model's improvement.} This toy experiment compares the performance contribution of fully labeled (a) and partially (b, w/o floor) labeled scans on S3DIS \cite{armeni2016s3dis} dataset. Specifically, the training dataset contains only 4 fully-labeled point cloud scans at the beginning. Another 4 fully or partially labeled scans are then added into the dataset at each following iteration. As shown in (c), compared to using all labels (solid line), removing floor labels (dash line) leads to similar performance on all classes including floor (blue), chairs (red), and bookcases (green). Additionally, (d) demonstrate that 12\% of point annotation (21.7M fully labeled points versus 19.1M partially labeled points at 20 scans) is saved by simply removing the floor labels. Therefore, this shows that not all annotated regions contribute to the model's improvement, and we can save the annotation costs by selecting key regions to annotate while maintaining the original performance.
    }
    \label{fig:fig2}
    \vspace{-1em}
\end{figure*}

To alleviate the huge burden of manual point-by-point labeling in large-scale point cloud datasets, some previous works have tried to reduce the total number of labeled point cloud scans \cite{lin2020efficient} or lower the annotation density within a single point cloud scan \cite{xu2020weaksup}. However, they neglect that regions in a point cloud scan may not contribute to the performance equally. As can be observed from Figure \ref{fig:fig2}, for a deep learning model, only 4 labeled point cloud scans are needed to reach over $0.9$ IoU on large uniform objects, such as floors. However, 20 labeled scans are required to achieve $0.5$ IoU on small items or objects with complex shapes and colors, like chairs and bookcases. Therefore, we argue that an effective point selection is essential for lowering annotation costs while preserving model performance.

In this work, we propose a novel \textit{Region-based and Diversity-aware Active Learning} (ReDAL) framework general for many deep learning network architectures. By actively selecting data from a huge unlabeled dataset for label acquisition, only a small portion of informative and diverse sub-scene regions is required to be labeled.

To find out the most informative regions for label acquisition, we utilize the combination of the three terms, \textit{softmax entropy}, \textit{color discontinuity}, and \textit{structural complexity}, to calculate the information score of each region. Softmax entropy is a widely used approach to measure model uncertainty, and areas with large color differences or complex structures in a point cloud provide more information because semantic labels are usually not smooth in these areas. As shown in the comparison of Figure \ref{fig:fig1} (a, b), the region-based active selection strategy significantly reduces the annotation effort of original full scene labeling.

Furthermore, to avoid redundant annotation resulting from multiple individually informative but duplicated data in a query batch, which is a common problem in deep active learning, we develop a novel diversity-aware selection algorithm considering both region information and diversity. In our proposed method, we first extract all regions' features, then measure the similarity between regions in the feature space, and finally, use a greedy algorithm to penalize multiple similar regions appearing in the same querying batch. As can be observed from the comparison of Figure \ref{fig:fig1} (b, c), our region-based and diversity-aware selection strategy can avoid querying labels for similar regions and further reduce the effort of manual labeling.

Experimental results demonstrate that our proposed method significantly outperforms existing deep active learning approaches on both indoor and outdoor datasets with various network architectures. On S3DIS \cite{armeni2016s3dis} and SemanticKITTI \cite{behley2019semantickitti} datasets, our proposed method can achieve the performance of 90\% fully supervised learning, while less than 15\%, 5\% annotations are required. Our ablation studies also verify the effectiveness of each component in our proposed method.

To sum up, our contributions are highlighted as follows,
\begin{itemize}
    \itemsep=2pt
    \item We pave a new path for 3D deep active learning that utilizes region segmentation as the basic query unit. 
    \item We design a novel diversity-aware active selection approach to avoid redundant annotations effectively.
    \item Experimental results show that our method can highly reduce human annotation effort on different state-of-the-art deep learning networks and datasets, and outperforms existing deep active learning methods.
\end{itemize}

\section{Related Work}

\subsection{Point Cloud Semantic Segmentation with less labeled data}

In the past decade, many supervised point cloud semantic segmentation approaches have been proposed \cite{lawin2017deep, qi2017pointnet, qi2017pointnet++, wang2019graph, thomas2019kpconv, choy2019minkowski, pcnn, liu2019pvcnn, tang2020spvconv}. However, despite the continuous development of supervised learning algorithms and the simplicity of collecting 3D point cloud data in large scenes, the cost of obtaining manual point-by-point marking is still high. As a result, many researchers began to study how to achieve similar performance with less labeled data.

Some have tried to apply transfer learning to this task. Wu \etal~\cite{wu2019squeezesegv2} developed an unsupervised domain adaptation approach to make the model perform well in real-world scenarios given only synthetic training sets. However, their method can only be applied to a single network architecture \cite{wu2018squeezeseg} instead of a general framework. 

Some others applied weakly supervised learning to reduce the cost of labeling. \cite{xu2020weaksup} utilized gradient approximation along with spatial and color smoothness constraints for training with few labeled scattered points. However, this operation does not save much cost, since annotators still have to switch viewpoints or zoom in and out throughout a scene when labeling scattered points. Besides, \cite{wei2020multi} designed a multi-path region mining module to help the classification model learn local cues and to generate pseudo point-wise labels at subcloud-level, but their performance is still far from the current state-of-the-art method compared with the fully-supervised result.

Still some others leveraged active learning to alleviate the annotation burden. \cite{luo2018mrf} designed an active learning approach to reduce the workload of CRF-based semantic labeling. However, their method can not be applied to current large-scale datasets for two reasons. First, the algorithm highly relies on the result of over-segmentation preprocessing and the algorithm cannot perfectly cut out small blocks with high purity in the increasingly complex scenes of the current real-world datasets. Second, the computation of pair-wise CRF is extremely high and thus not suitable for large-scale datasets. In addition to the above practice, \cite{lin2020efficient} proposed segment entropy to measure the informativeness of single point cloud scan in a deep active learning pipeline. 

To the best of our knowledge, we are the first to design a region-based active learning framework general for many deep learning models. Furthermore, our idea of reducing redundant annotation through  diversity-aware selection is totally different from those previous works.

\subsection{Deep Active Learning}
Sufficient labeled training data is vital for supervised deep learning models, but the cost of manual annotation is often high.

Active Learning  \cite{settles2009active} aims to reduce labeling cost by selecting the most valuable data for label acquisition. \cite{ijcnn} proposed the first active learning framework on deep learning where a batch of items, rather than a single sample in traditional active learning, is queried in each active selection for acceleration.
Several past deep active learning practices are based on model uncertainty. \cite{ijcnn} is the first work that applied least confidence, smallest margin \cite{roy2001toward} and maximum entropy \cite{shannon1948mathematical} to deep active learning. \cite{wang2016cost} introduced semi-supervision to active learning, which assigned pseudo-labels for instances with the highest certainty. \cite{gal2016dropout, gal2017deep} combined bayesian active learning with deep learning, which estimated model uncertainty by MC-Dropout.

In addition to model uncertainty, many recent deep active learning works take in-batch data diversity into account. \cite{sener2018coreset, kirsch2019batchbald, ash2020babdge} stated that neglecting data correlation would cause similar items to appear in the same querying batch, which further leads to inefficient training. 
\cite{sener2018coreset} converted batch selection into a core-set construction problem to ensure diversity in labeled data; \cite{kirsch2019batchbald, ash2020babdge} tried to consider model uncertainty and data diversity at the same time. 
Empirically, uncertainty and diversity are two key indicators of active learning. \cite{hsu2015albl} is a hybrid method that enjoys the benefit of both by dynamically choosing the best query strategy in each active selection step.

To the best of our knowledge, we design the first 3D deep active learning framework combining uncertainty, diversity and point cloud domain knowledge.


\section{Method}

\begin{figure*}[t!]
    \centering
    \includegraphics[width=\textwidth]{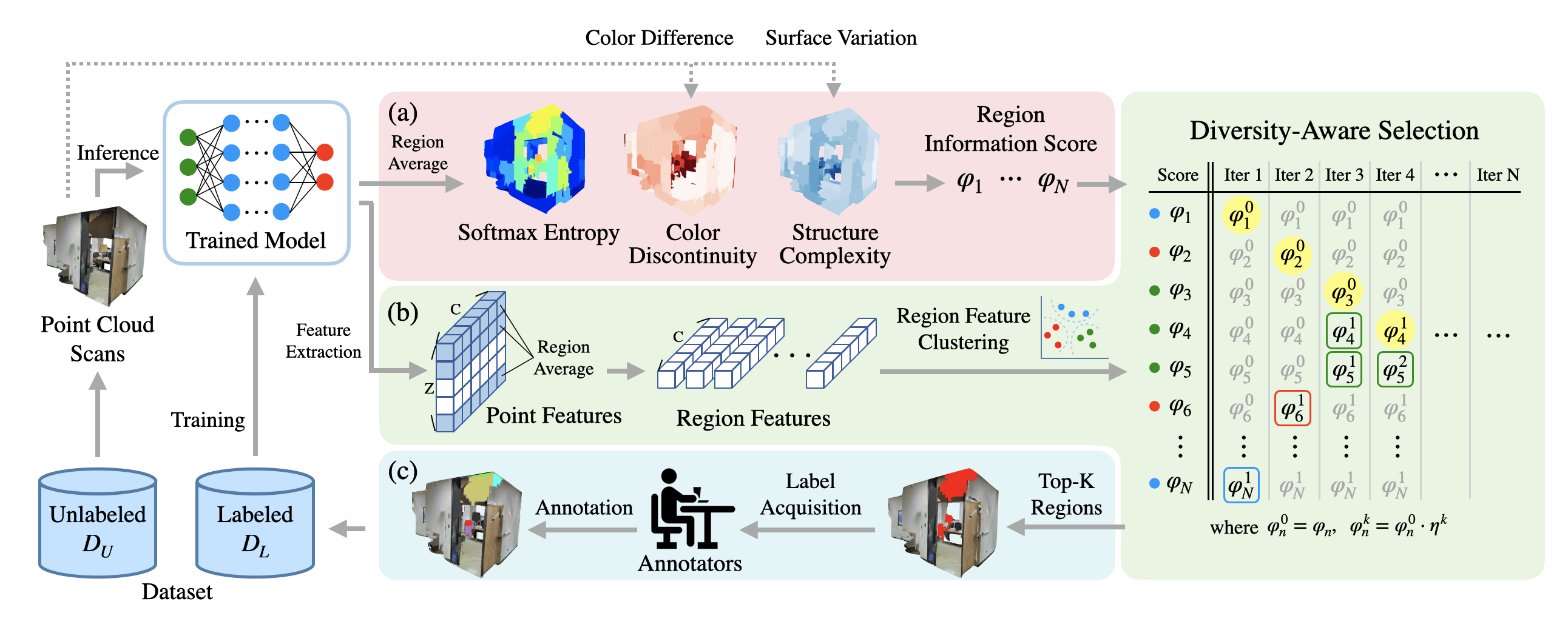}
    \caption{\textbf{Region-based and Diversity-Aware Active Learning Pipeline.} In the proposed framework, a point cloud semantic segmentation model is first trained in supervision with labeled dataset $D_L$. The model then produces softmax entropy and features of all regions from the unlabeled dataset $D_U$. (a) Softmax entropy along with color discontinuity and structure complexity calculated from the unlabeled regions serves as selection indicators (Sec. \ref{subsec:method_informative}), and (b) generates scores which are then adjusted by penalizing regions belonging to the same clusters grouped by the extracted features (Sec. \ref{subsec:importance}). (c) The top-ranked regions are labeled by annotators and added to the labeled dataset $D_L$ for the next phase (Sec. \ref{subsec:label}).
    }
    \label{fig:AL_pipeline}
\end{figure*}

\label{sec:method}

In this section, we describe our region-based and diversity-aware active learning pipeline in detail. Initially, we have a 3D point cloud dataset $D$, which can be divided into two parts. One is a subset $D_L$ containing randomly selected point cloud scans with complete annotations, and the other is a large unlabeled set $D_U$ without any annotation. 

In traditional deep active learning, the network is trained on the current labeled set $D_L$ under supervision initially. Then, select a batch of data for label acquisition from the unlabeled set $D_U$ according to a certain strategy. Finally, move the newly labeled data from $D_U$ to $D_L$; then, go back to step one to re-train or fine-tune the network and repeat the loop until the budget of the annotation is exhausted.

\subsection{Overview}

We use a sub-scene region as the fundamental query unit in our proposed ReDAL method. In traditional deep active learning, the smallest unit for label querying is a sample, which is a whole point cloud scan in our task. However, based on the prior experiment shown in Figure \ref{fig:fig2}, we know that some labeled regions may contribute little to the model improvement. Therefore, we change the fundamental unit of label querying from a point cloud scan to a sub-scene region in a scan.

Instead of using model uncertainty as the only criterion to determine the selection common in 2D active learning, we leverage the domain knowledge from 3D computer vision and include two informative cues, \textit{color discontinuity} and \textit{structural complexity}, in the selection indicators. Moreover, to avoid redundant labeling caused by multiple duplicate regions in a querying batch, we design a simple yet effective diversity-aware selection strategy to mitigate the problem and improve the performance.

Our region-based and diversity-aware active learning can be divided into 4 steps: (1) Train on current labeled dataset $D_L$ in a supervised manner. (2) Calculate the region information score $\varphi$ for each region with three indicators: softmax entropy, structure complexity and color discontinuity as shown in Figure \ref{fig:AL_pipeline} (a) (Sec. \ref{subsec:method_informative}). (3) Perform diversity-aware selection by measuring the similarity between all regions and using a greedy algorithm to penalize similar regions appearing in a querying batch as shown in Figure \ref{fig:AL_pipeline} (b) (Sec. \ref{subsec:importance}). (4) Select top-K regions for label acquisition, and move them from the unlabeled dataset $D_U$ into the current labeled dataset $D_L$ as shown in Figure \ref{fig:AL_pipeline} (c) (Sec. \ref{subsec:label}).

\subsection{Region Information Estimation} \label{subsec:method_informative}

We divide a large-scale point cloud scan into some sub-scene regions as the fundamental label querying units using VCCS \cite{papon2013vccs} algorithm, an unsupervised over-segmentation method that groups similar points into a region. The original purpose of this algorithm was to cut a point cloud into multiple small regions with high segmentation purity to reduce the computational burden of the probability statistical model. Different from the original purpose of requiring high purity, our method merely utilizes the algorithm to divide a scan into sub-scenes of median size for better annotation and learning. An ideal sub-scene consists of several but not complicated semantic meanings, while preserving geometric structures of point cloud.

In each active selection step, we calculate the information score of a region from three aspects: (1) softmax entropy, (2) color discontinuity, and (3) structural complexity, which is described in detail as follows.
\subsubsection{Softmax Entropy}
Softmax entropy is a widely used approach to measure the uncertainty in active learning \cite{ijcnn, wang2016cost}. We first obtain the softmax probability of all point cloud scans in the unlabeled set $D_U$ with the model trained in the previous active learning phase. Then, given the softmax probability $P$ of a point cloud scan, we calculate the region entropy $H_n$ for the $n$-th region $R_n$ by averaging the entropy of points belonging to the region $R_n$ as shown in Eq. \ref{eq:entropy}.

\begin{equation}
    H_n = \frac{1}{|R_n|}\sum_{i \in R_n} -P_i \log P_i
    \label{eq:entropy}
\end{equation}

\subsubsection{Color Discontinuity}
In 3D computer vision, the color difference is also an important clue since areas with large color differences are more likely to indicate semantic discontinuity. Therefore, it is also included as an indicator for measuring regional information. For all points in a given point cloud with color intensity value $I$, we compute the $1$-norm color difference between a point $i$ and its $k$-nearest neighbor $d_i$ ($|d_i| = k$). Then we produce the region color discontinuity score $C_n$ for the $n$-th region $R_n$ by averaging the values of points belonging to the region $R_n$ as shown in Eq. \ref{eq:color}. 

\begin{equation}
    C_n = \frac{1}{k\cdot |R_n|}\sum_{i \in R_n} \sum_{j \in d_i} || I_i - I_j ||_1
    \label{eq:color}
\end{equation}

\subsubsection{Structural Complexity}

We also include structure complexity as an indicator, since complex surface regions, boundary places, or corners in a point cloud are more likely to indicate semantic discontinuity.  
For all points in a given point cloud, we first compute the surface variation $\sigma$ based on \cite{bazazian2015curvature, pauly2003surface}. Then, we calculate the region structure complexity score $S_n$ for the $n$-th region $R_n$ by averaging surface variation of points belonging to the region $R_n$ as shown in Eq. \ref{eq:structure}.

\begin{equation}
    S_n = \frac{1}{ |R_n|}\sum_{i \in R_n} \sigma_i
    \label{eq:structure}
\end{equation}

After calculating the softmax entropy, color discontinuity, and structural complexity of each region, we combine these terms linearly to form the region information score $\varphi_n$ of the $n$-th region $R_n$ as shown in Eq. \ref{eq:combine}.

\begin{equation}
    \varphi_n = \alpha H_n + \beta C_n + \gamma S_n
    \label{eq:combine}
\end{equation}

Finally, we rank all regions in descending order based on the region information scores and produce a sorted information list $\varphi = (\varphi_1, \varphi_2, \cdots,  \varphi_N)$. The above process is illustrated in Figure \ref{fig:AL_pipeline} (a).

\subsection{Diversity-aware Selection} \label{subsec:importance}
\begin{figure}
    \centering
    \includegraphics[width=\linewidth]{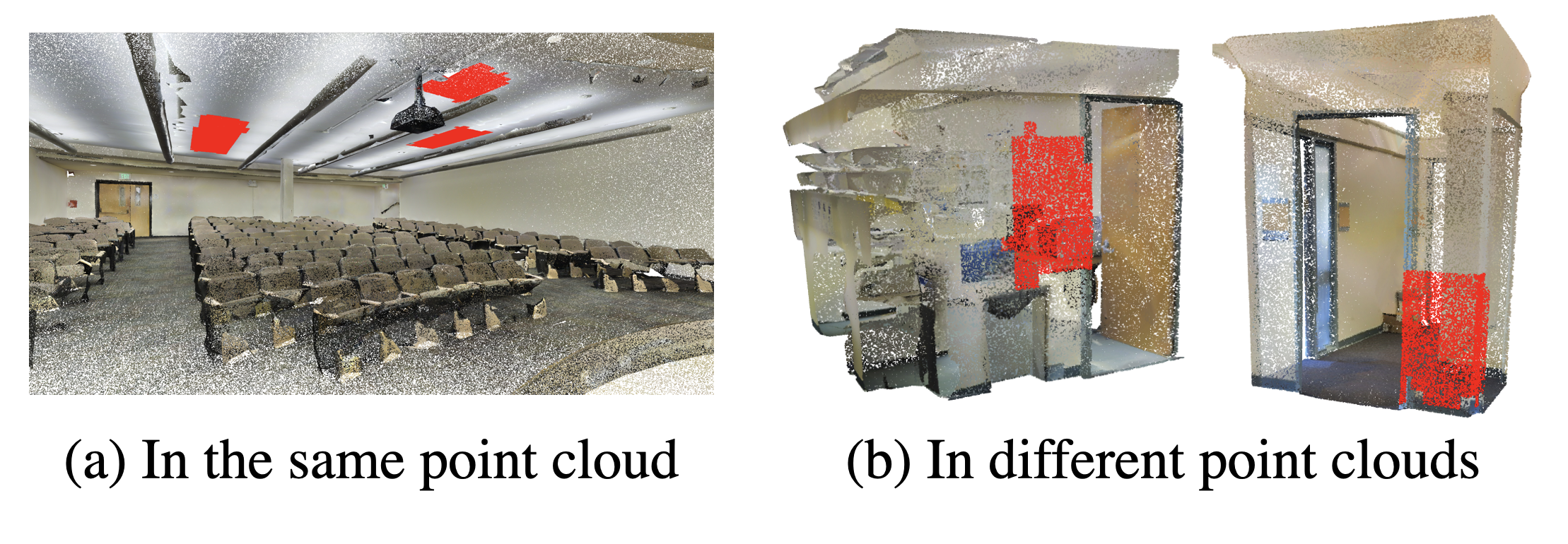}
    \caption{\textbf{Our method is able to find out visually similar regions not only in the same point cloud (a) but also in different point clouds (b).} The areas colored in red are the ceiling in an auditorium (a) and walls next to the door (b).
    These regions may cause redundant labeling effort if appearing in the same querying batch, and thus they are filtered by our diversity-aware selection (Sec. \ref{subsubsec:similarity}).
    }
    \label{fig:vis_sim}
\end{figure}

With the sorted region information list $\varphi$, a naive way is to select the top-ranked regions for label acquisition directly. Nevertheless, this strategy results in multiple visually similar regions being in the same batch as shown in Figure \ref{fig:vis_sim}. These regions, though informative individually, provide less diverse information for the model. 

To avoid visually similar regions appearing in a querying batch, we design a diversity-aware selection algorithm divided into two parts: (1) region similarity measurement and (2) similar region penalization.

\subsubsection{Region Similarity Measurement}
\label{subsubsec:similarity}
We measure the similarity among regions in the feature space rather than directly on point cloud data because the scale, shape, and color of each region are totally different.

Given a point cloud scan with $Z$ points, we record the output before the final classification layer as the point features with shape $Z \times C$. Then, we produce the region features by calculating the average of the point features of the points belonging to the same region. Finally, we gather all point cloud regions and use $k$-means algorithm to cluster these region features. The above process can be seen in the middle of Figure \ref{fig:AL_pipeline} (b). After clustering regions, we regard regions belonging to the same cluster as similar regions. An example is shown in Figure \ref{fig:vis_sim}.

\subsubsection{Similar Region Penalization}

To select diverse regions, a greedy algorithm takes the sorted list of information scores $\varphi \in \mathbb{R}^N$ as input and re-scores all regions by penalizing regions with lower scores that belong to the same clusters containing regions with higher scores. 

The table in the right of Figure \ref{fig:AL_pipeline} (b) offers an example where the algorithm loops through all regions one by one. The scores of regions ranked below while belonging to the same cluster as the current region are multiplied by a decay rate $\eta$. To be specific, the red, green, and blue dots in the left of the table denote the cluster indices of regions, and $\varphi_n^k$ denotes the score $\varphi_n$ of the region $R_n$ with $k$-time penalization. Yellow circles under $\varphi_n^k$ indicate the current region to be compared in each iteration, and rounded rectangles mark regions belonging to the same cluster as the current region. In the first iteration, $\varphi_N$ is penalized as $R_N$ and $R_0$ belong to the same cluster denoted by blue dots. $R_N$'s score is then replaced by $\varphi_N^1$ to mark the first decay. In the third iteration, $\varphi_4$ and $\varphi_5$ are both penalized as $R_3$, $R_4$ and $R_5$ belong to the same cluster denoted by green dots. Their scores are thus substituted by $\varphi_4^1$ and $\varphi_5^1$. The same logic applies to the other iterations. Then we obtain the adjusted scores $\varphi_N^*$ for label acquisition.

Note that in our implementation shown in Algorithm $\ref{algo:penalty}$, we penalize the corresponding importance weight $W$, which are initialized to $1$ for all $M$ clusters, instead of directly penalizing the score for efficiency. Precisely, in each iteration, we adjust the score of the pilot by multiplying the importance weight of its cluster. Then, the importance weight of the cluster is multiplied by the decay rate $\eta$.

\begin{algorithm}

\SetAlgoLined
\KwIn{Original sorted information score $\varphi \in \mathbb{R}^N$ and corresponding $M$-cluster region labels $L \in \mathbb{R}^N$; cluster importance weight $W \in \mathbb{R}^M$ and decay rate $\eta$}
\KwOut{Final Region information score $\varphi^* \in \mathbb{R}^N$}
$\textbf{Init: } W_m\leftarrow 1 \quad  \forall 1 \leq m \leq M$\;
 \For{$i \leftarrow 1$ to $N$}{
    $\varphi_i^{*} \leftarrow \varphi_i \cdot W_{L_i}$\;
    $W_{L_i} \leftarrow W_{L_i} \cdot \eta$\; 
 }
 \Return $\varphi^*$
 \caption{Similar Region Penalization}
 \label{algo:penalty}
\end{algorithm}

\subsection{Region Label Acquisition} \label{subsec:label}

After getting the final scores $\varphi^*$ by considering region diversity, we select regions into a querying batch in decreasing order according to $\varphi^*$ for label acquisition until the budget of this round is exhausted. Note that in each label acquisition step, we set the budget as a fixed number of total points instead of a fixed number of regions for fair comparison since each region contains different number of points.

For experiments, after selecting the querying batch data, we regard the ground truth region annotation as the labeled data obtained from human annotators. Then, these regions are moved from unlabeled set $D_U$ to labeled set $D_L$.
Note that different from the 100\% fully labeled initial training point cloud scans, since we regard a region as the basic labeling unit, many point cloud scans with only a small portion of labeled region are appended to the labeling data set $D_L$ in every active selection step as shown in Figure \ref{fig:AL_pipeline} (c).

Finishing the active selection step containing region information estimation, diversity-aware selection and region label acquisition, we repeat the active learning loop to fine-tune the network on the updated labeled dataset $D_L$.

\section{Experiments}

\begin{figure*}[t!]
    \centering
    \includegraphics[width=\textwidth]{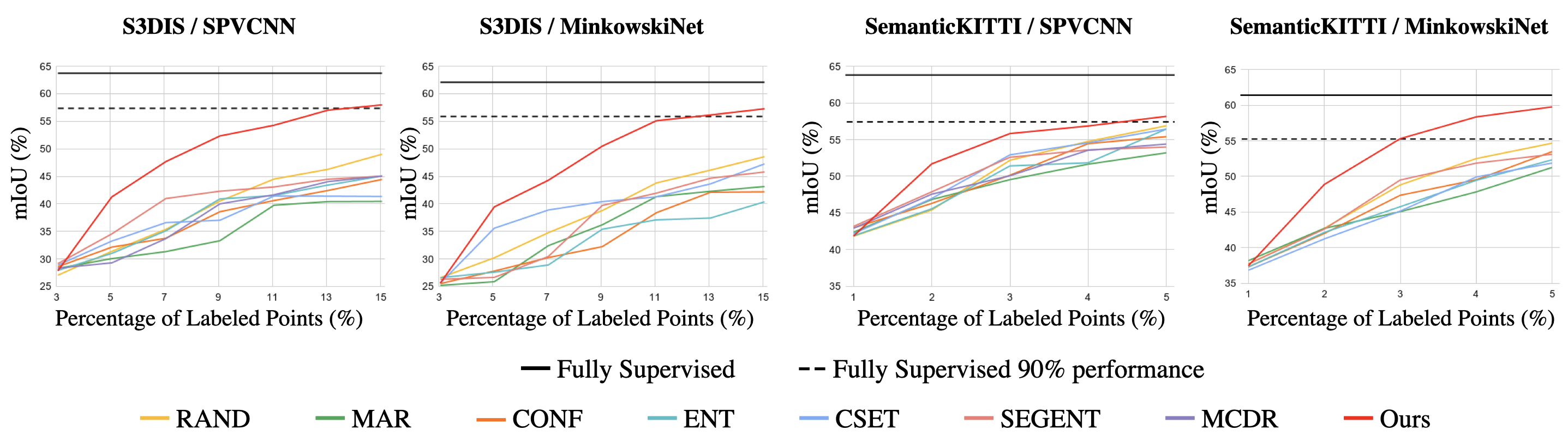}
    \caption{\textbf{Experimental results of different active learning strategies on 2 datasets and 2 network architectures.} We compare our region-based and diversity-aware active selection strategy with other existing baselines. It is obvious that our proposed method outperforms any existing active selection approaches under any combinations. Furthermore, our method is able to reach 90\% fully supervised result with only 15\%, 5\% labeled points on S3DIS \cite{armeni2016s3dis} and SemanticKITTI \cite{behley2019semantickitti} dataset respectively.
    }
    \label{fig:performance}
\end{figure*}

\subsection{Experimental Settings} \label{subsec:exp_setting}

In order to verify the effectiveness and universality of our proposed active selection strategy, we conduct experiments on two different large-scale datasets and two different network architectures. The implementation details are explained in the supplementary material due to limited space.

\paragraph{Datasets.} 
We use S3DIS \cite{armeni2016s3dis} and SemanticKITTI \cite{behley2019semantickitti} as representatives of indoor and outdoor scenes, respectively. S3DIS is a commonly used indoor scene segmentation dataset. The dataset can be divided into 6 large areas, with a total of 271 rooms. Each room has a corresponding dense point cloud with color and position information. We evaluate the performance of all label acquisition strategies on the Area5 validation set and perform active learning on the remaining datasets. SemanticKITTI is a large-scale autonomous driving dataset with 43552 point cloud scans from 22 sequences. Each point cloud scan is captured by LiDAR sensors with only position information. We evaluate the performance of all label acquisition strategies on the official validation split (seq 08) and perform active learning on the whole official training split (seq 00$\sim$07 and 09$\sim$10).

\paragraph{Network Architectures.} 

To verify an active strategy on various deep learning networks, we use MinkowskiNet \cite{choy2019minkowski}, based on sparse convolution, and SPVCNN \cite{tang2020spvconv}, based on point-voxel CNN thanks to the great performance on large-scale point cloud datasets with high inference speed.

\paragraph{Active Learning Protocol.}

For all experiments, we first randomly select a small portion ($x_{init}$\%) of fully labeled point cloud scans from the whole training data as the initial labeled data set $D_L$ and treat the rest as the unlabeled set $D_U$. Then, we perform $K$ rounds of the following actions: (1) Train the deep learning model on $D_L$ in a supervised manner. (2) Select a small portion ($x_{active}$\%) of data from $D_U$ for label acquisition according to different active selection strategies. (3) Add the newly labeled data into $D_L$ and finetune the deep learning model.

We choose $x_{init} = 3\%$, $K = 7$, and $x_{active} = 2\%$ for S3DIS dataset, and $x_{init} = 1\%$, $K = 5$, and $x_{active} = 1\%$ for SemanticKITTI dataset. To ensure the reliability of the experimental results, we perform the experiments three times and record the average value for each setting.

\subsection{Comparison among different active selection strategies.}

We compare our proposed method with 7 other active selection strategies, including random point cloud scans selection (\textbf{RAND}), softmax confidence (\textbf{CONF}) \cite{ijcnn}, softmax margin (\textbf{MAR}) \cite{ijcnn}, softmax entropy  (\textbf{ENT}) \cite{ijcnn}, MC-dropout (\textbf{MCDR}) \cite{gal2016dropout, gal2017deep}, core-set approach, (\textbf{Core-Set}) \cite{sener2018coreset} and segment-entropy  (\textbf{SEGENT}) \cite{lin2020efficient}. The implementation details are explained in the supplementary material.

The experimental results can be seen in Figure \ref{fig:performance}. In each subplot, the x-axis means the percentage of labeled points and the y-axis indicates the mIoU achieved by the network. Our proposed ReDAL significantly surpasses other existing active learning strategies under any combination. 

In addition, we observe that random selection (\textbf{RAND}) outperforms any other active learning methods except ours on four experiments. For uncertainty-based methods, such as ENT and MCDR, since the model uncertainty value is dominated by background area, the performance is not as expected. The same, for pure diversity approach, such as CSET, since the global feature is dominated by the background area, simply clustering global feature cannot produce diverse label acquisition. The experimental results further verify our proposal of changing the fundamental querying units from a scan to a region is a better choice.

On the S3DIS \cite{armeni2016s3dis} dataset, our proposed active selection strategy can achieve more than 55 \% mIoU with 15 \% labeling points, while others cannot reach 50 \% mIoU under the same condition. The main reason for such a large performance gap is that these room-sized point cloud data in the dataset is very different. Compared with other active selection methods querying a batch of point cloud scans, our region-based label acquisition  make the model be trained on more diverse labeled data.

As for the SemanticKITTI \cite{behley2019semantickitti}, we find that with merely less than 5\% of labeled data, our active learning strategy can achieve 90\% of the result of fully supervised methods. With the network architecture of MinkowskiNet, our active selection strategy can even reach 95 \% fully supervised result with only 4 \% labeled points.

Furthermore, in Table \ref{tab:perclass_comparison}, the performance of some small or complicated class objects like bicycles and bicyclists is even better than fully supervised one. Table~\ref{tab:perclass_sample} shows the reason that our selected algorithm focuses more on those small or complicated objects. In other words, our ReDAL does not waste annotation budgets on easy cases like uniform surfaces, which again realizes the observation and motivation in the introduction. Besides, our ReDAL selection strategy makes it more friendly to real-world applications, such as autonomous driving, since it puts more emphasis on important and valuable semantics.

\begin{table}
    \centering
    \begin{tabular}{l||c||ccccc}
         Method & avg & road & person & bicycle & bicyclist   \\ \hline \hline
         RAND & 54.7  & 90.2 & 52.0 &  9.5 & 47.7  \\
         Full  & \textbf{61.4} & \textbf{93.5} & \textbf{65.0} & 20.3 & 78.4   \\
         \textbf{ReDAL}  & 59.8 & 91.5 & 63.4 & \textbf{29.5} & \textbf{84.1}  \\
    \end{tabular}
    \caption{
     \textbf{Results of IoU performance (\%) on SemanticKITTI~\cite{behley2019semantickitti}.}
     Under only 5 \% of annotated points, our proposed ReDAL outperforms random selection and is on par with full supervision (Full).}
    \label{tab:perclass_comparison}
\end{table}

\begin{table}[t!]
    \centering
    \begin{tabular}{l||ccccc}
         Method & road & person & bicycle & bicyclist   \\ \hline \hline
         RAND & 206  & 0.42  &  0.15 & 0.10  \\
         Full & 205  & 0.35  & 0.17 & 0.13  \\
         \textbf{ReDAL} & 168 & 1.20 & 0.25 & 0.21   \\
    \end{tabular}
    \caption{
    \textbf{Labeled Class Distribution Ratio (\textperthousand).} With limited annotation budgets, our active method ReDAL queries more labels on small objects like a person but less on large uniform areas like roads. The selection strategy can mitigate the label imbalance problem and improve the performance on more complicated object scenes without hurting much on large areas as shown in Table~\ref{tab:perclass_comparison}.
    } 
    
    \label{tab:perclass_sample}
\end{table}

\begin{figure*}
    \centering
    \includegraphics[width=\textwidth]{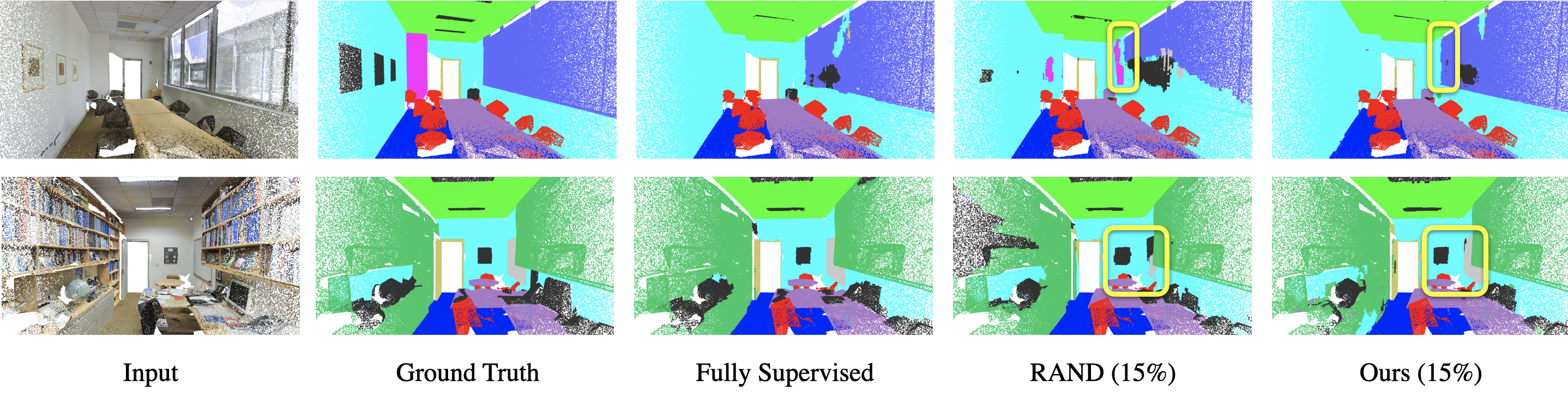}
    \caption{\textbf{Visualization for the inference result on S3DIS dataset with SPVCNN network architecture.} We show some inference examples on S3DIS Area 5 validation set. With our active learning strategy, the model can produce sharp boundaries (shown on the yellow bounding box in the first row) and recognize small objects, such as boards and chairs (shown on the yellow bounding box in the second row) with only 15 \% labeled points.
    }
    \label{fig:s3dis_vis}
\end{figure*}

\begin{figure*}
    \centering
    \includegraphics[width=\textwidth]{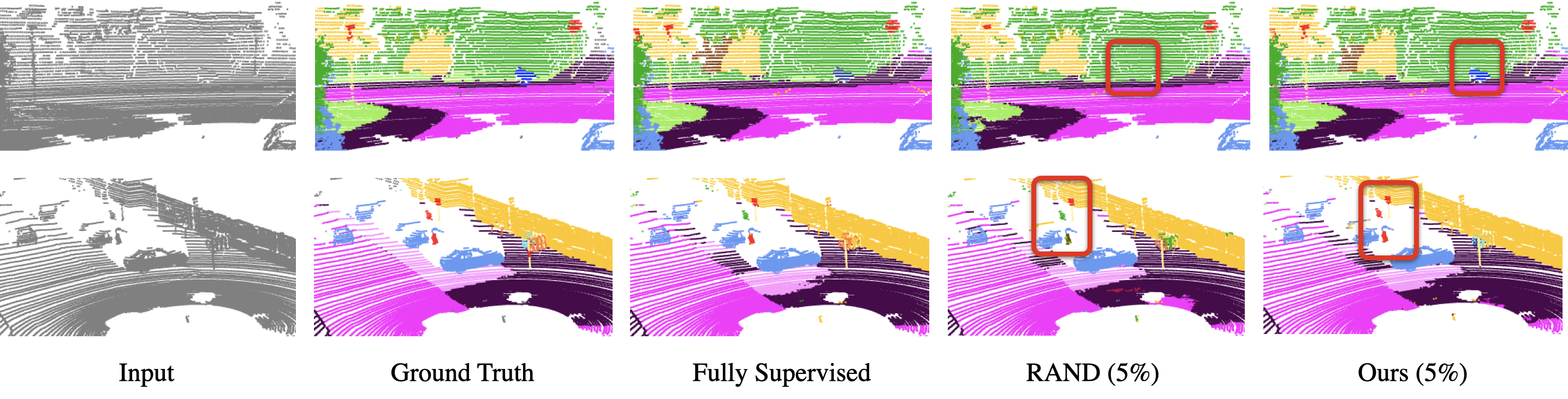}
    \caption{\textbf{Visualization for the inference result on SemanticKITTI dataset with MinkowskiNet network architecture.} We show some inference examples on the SemanticKITTI sequence 08 validation set. With our active learning strategy, the model can correctly recognize small vehicles (shown on the red bounding box in the first row) and identify people on the side walk (shown on the red bounding box in the second row) with merely 5 \% labeled points.
    }
    \label{fig:semkitti_vis}
\end{figure*}

\begin{figure}[t]
    \centering
    \includegraphics[width=\linewidth]{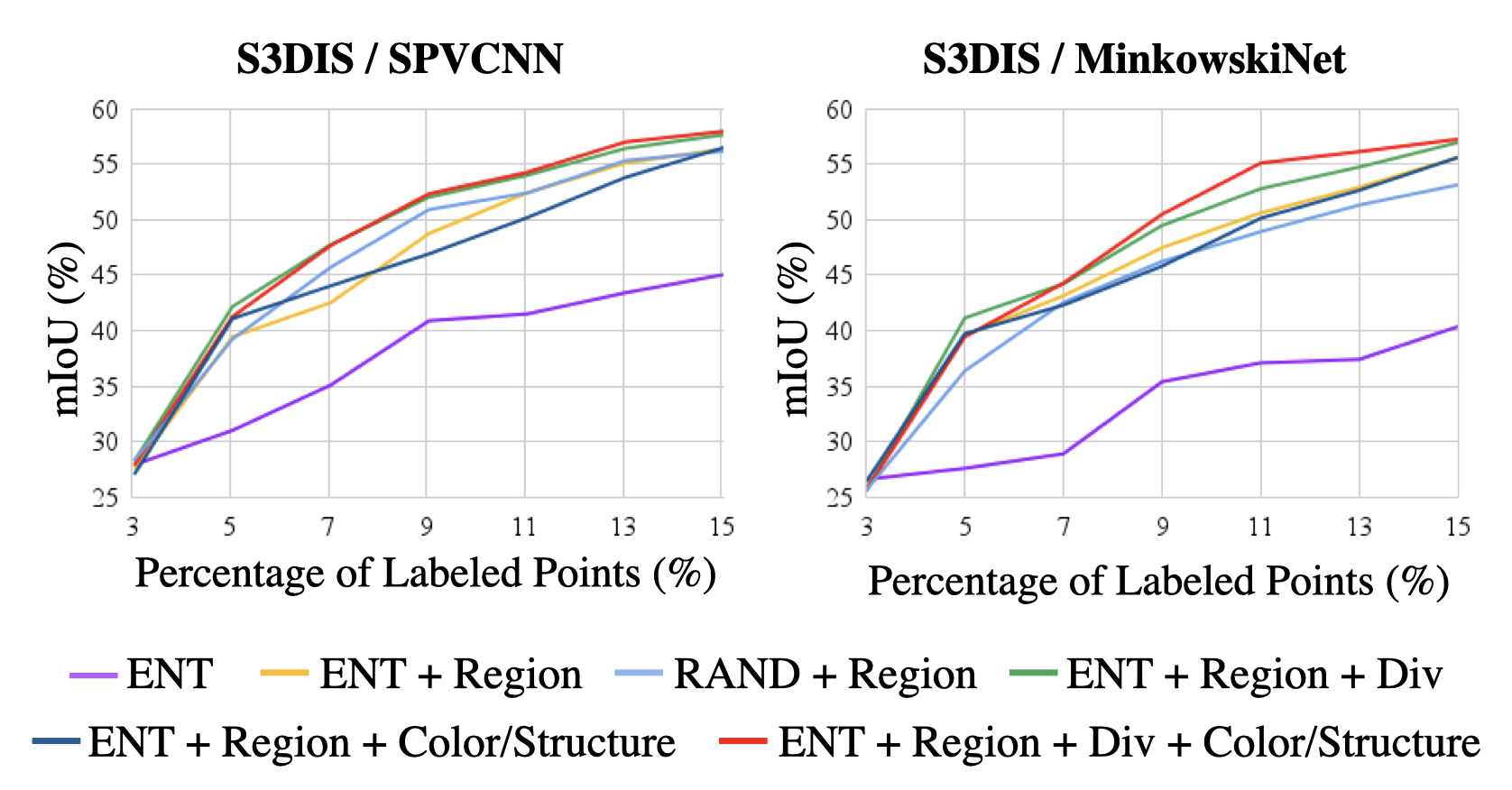}
    \caption{
    \textbf{Ablation Study.} The best combinations are altering labeling units from scans to regions (+Region), applying diversity-aware selection (+Div), and additional region information (+Color/Structure). Best viewed in color. (Sec. \ref{subsec:ablation})
    }
    \label{fig:ablation}
\end{figure}

\subsection{Ablation Studies}
\label{subsec:ablation}

We verify the effectiveness of all components in our proposed active selection strategy on S3DIS dataset \cite{armeni2016s3dis}. The results are shown in Figure \ref{fig:ablation}.

First, changing the labeling units from scans to regions contributes the most to the improvement as shown from the comparison of the purple line (ENT), the yellow line (ENT+Region), and the light blue line (RAND+Region). By applying region-based selection, the mIoU performance improves over 10 \% under both networks.

Furthermore, our diversity-aware selection also plays a key role in the active selection process shown from the comparison of the yellow line (ENT+Region) and the green line (ENT+Region+Div). Without the aid of this component, the performance of region-based entropy is even lower than random region selection under SPVCNN network architecture as shown from the comparison of the yellow line (ENT+Region) and the light blue line (RAND+Region).

As for adding extra information of color discontinuity and structure complexity, it contributes little to SPVCNN but is helpful to MinkowskiNet when the percentage of labeled points is larger than 9 \% as shown in the comparison of the green line (w/o color and structure) and the red line (w/ color and structure). 

Note that the performance of ``ENT+Region" (the yellow line) is similar to ``ENT+Region+Color/Structure" (the dark blue line). The reason is that without our diversity module, the selected query batch is still full of duplicated regions. The result also validates the importance of our diversity-aware greedy algorithm.

\section{Conclusion}

We propose ReDAL, a region-based and diversity-aware active learning framework, for point cloud semantic segmentation. The active selection strategy considers region information and diversity, concentrating the labeling effort on the most informative and distinctive regions rather than full scenes. This approach can be applied to many deep learning network architectures and datasets, substantially reducing the annotation cost, and greatly outperforms existing active learning strategies.

\section*{Acknowledgement}
This work was supported in part by the Ministry of Science and Technology, Taiwan, under Grant MOST 110-2634-F-002-026,  Mobile Drive Technology (FIH Mobile Limited), and Industrial Technology Research Institute (ITRI). We benefit from NVIDIA DGX-1 AI Supercomputer and are grateful to the National Center for High-performance Computing.

{\small
\bibliographystyle{ieee_fullname}
\bibliography{egbib}
}

\clearpage

\appendix
\begin{center}
\large{\textbf{Supplementary Material}}
\end{center}

The supplementary material is organized as follows: Section \ref{sec:impl} describes the implementation details. Section \ref{sec:al} explains the baseline active learning methods. Section \ref{sec:exp} shows the original data of line charts or tables in the main paper.

\section{Implementation Details}
\label{sec:impl}

As explained in the main paper, the pipeline of our ReDAL contains four steps: (1) Train the deep learning model in supervision with labeled dataset $D_L$. (2) Calculate region information score using softmax entropy, color discontinuity, and structure complexity. (3) Diversity-aware selection by penalizing visually similar regions appearing in the same querying batch. (4) The top-ranked regions are labeled by annotators and added to the labeled dataset $D_L$. This section explains the implementation details of the first three steps, and the fourth step has been explained clearly in the main paper. Note that the following symbols are the same as those in Section 3 of the main paper.

\subsection{Network Training}

For both S3DIS \cite{armeni2016s3dis} and SemanticKITTI \cite{behley2019semantickitti} datasets, the networks are trained with Adam optimizer (initial learning rate = $0.001$) and cross-entropy loss. We train the network on 8 V100 GPUs and set the batch size to 16. We set voxel resolution to 5cm for both datasets.

On the S3DIS dataset, the deep learning model was trained for 200 epochs on 3\% of the initial fully labeled point cloud scan and then fine-tuned for 150 epochs after adding 2\% labeled data each time for both network architecture backbones. On the SemanticKITTI dataset, the deep learning model was trained for 100 epochs on 1\% of the initial fully labeled point cloud scan and then fine-tuned for 30 epochs after adding 1\% labeled data each time for both network architectures.

\subsection{Region Information Estimation}

\begin{figure}
    \centering
    \includegraphics[width=0.99\linewidth]{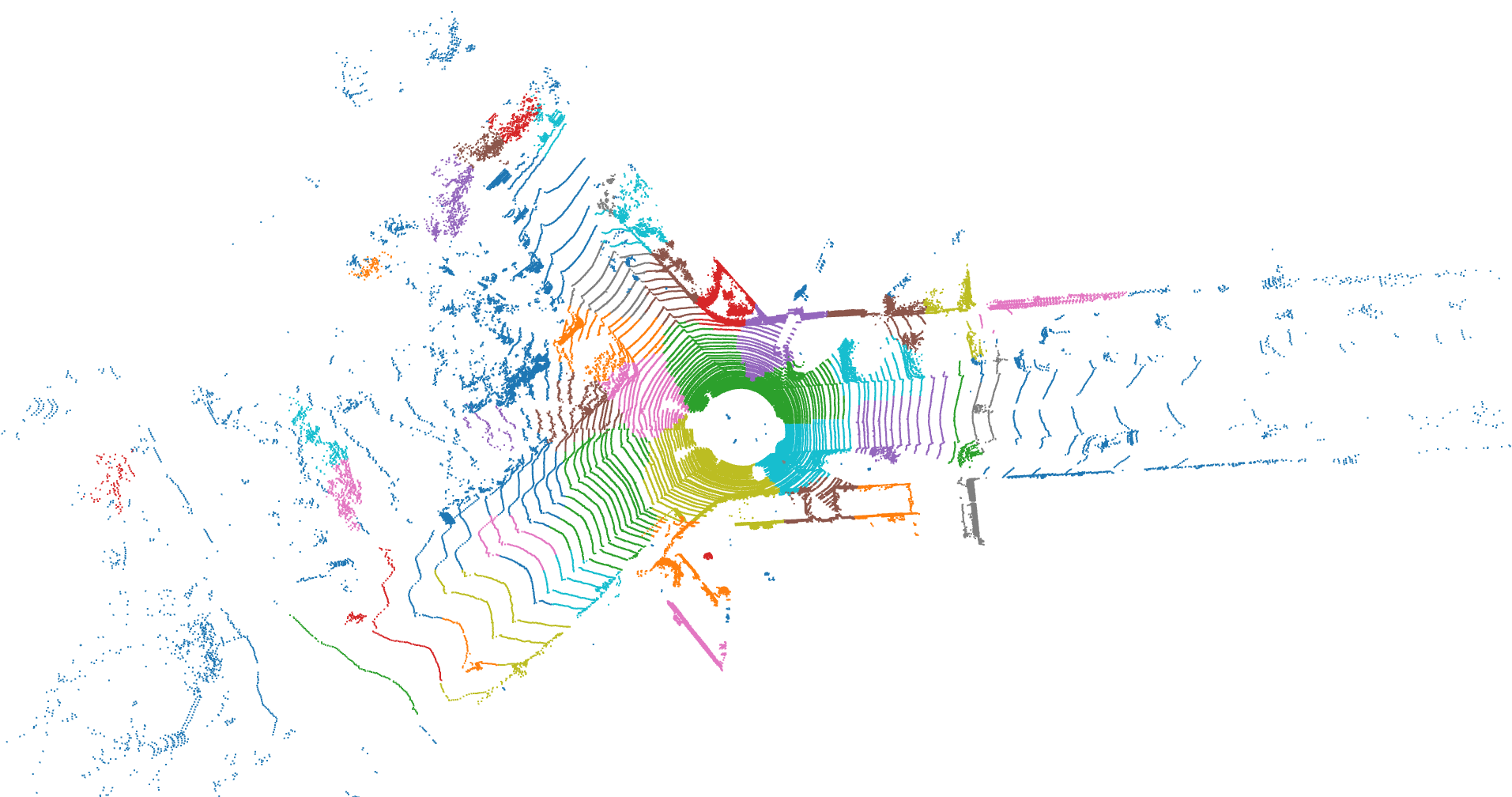}
    \caption{\textbf{Visualization of divided sub-scene regions in SemanticKITTI dataset.} Points of the same color in neighboring places belong to the same region.}
    \label{fig:semkitti_supvox}
\end{figure}

We utilize the VCCS algorithm \cite{papon2013vccs} to divide a 3D scene into multiple sub-scene regions. In the algorithm, the whole 3D space is initially divided into multiple regions with two hyper-parameters $R_{seed}, R_{voxel}$, where $R_{seed}$ indicates the initial distance between regions and $R_{voxel}$ represents the minimal region resolution. After that, the clustering procedure adjusts the region boundary based on spatial or color connectivity iteratively. For the S3DIS dataset, we set  $R_{seed}, R_{voxel}$ to a small value ($R_{seed} = 1.0, R_{voxel} = 0.1$) since objects in an indoor scene are small. For the SemanticKITTI dataset, we set $R_{seed}, R_{voxel}$ to a large value ($R_{seed} = 10, R_{voxel} = 0.5$). The reason is that the point cloud is sparse in outdoor 3D space, and choosing larger parameters ($R_{seed}, R_{voxel}$) can avoid creating small, unrepresentative regions. An example of divided sub-scene regions of the SemanticKITTI dataset is shown in Figure \ref{fig:semkitti_supvox}. 


As mentioned in Section 3.2 of the main paper, we linearly combine softmax entropy, color discontinuity, and structural complexity as region information score. For color discontinuity and structural complexity, we calculate color differences and surface variation for each point and its $k$-nearest neighbors ($k = 50$ in both datasets). As for the weight of the linear combination of these three terms, which is described in Eq. 4 of the main paper, we set $\alpha = 1, \beta = 0.1, \gamma = 0.05$ for S3DIS dataset and $\alpha = 1, \beta = 0, \gamma = 0.05$ for SemanticKITTI dataset. Note that the value $\alpha = 1.0, \beta = 0.1, \gamma = 0.05$ is empirically decided for we found that model uncertainty is much more important than the color discontinuity and structural complexity terms. In addition, since the SemanticKITTI dataset does not have point-by-point color information, we set $\beta = 0$ for the dataset.

\subsection{Diversity-aware Selection}

As explained in Section 3.3 of the main paper, we measure the similarity of these regions by clustering their corresponding region features. We set the number of clusters of all regions $M=400, 150$ for the S3DIS and SemanticKITTI datasets, respectively. For both datasets, we set the decay rate $\eta = 0.95$. Note that our diversity-aware selection algorithm does not create too much computational burden. On the SemanticKITTI dataset, our diversity-aware selection algorithm only takes only 0.58 ms per region on average.

Note that we empirically found that k in k-nn (mentioned in the previous sub-section), decay rate $\eta$ and the number of clusters $M$ is not sensitive to the experimental results, where all values are determined via grid search. 




\section{Baseline Active Learning Methods}
\label{sec:al}

In this section, we describe the implementation of the baseline active learning methods used in our experiments.

\paragraph{Random selection (RAND)} Randomly select a portion of point cloud scans in the unlabeled dataset for label acquisition. The strategy is commonly used as the baseline for active learning methods \cite{ijcnn, gal2017deep, sener2018coreset, lin2020efficient}.

\paragraph{Margin sampling (MAR)} Some previous active learning methods query instances with the smallest model decision margin, which is the predicted probability difference between the two most likely class labels \cite{ijcnn}. As shown in Eq. \ref{eq:mar}, given a point cloud scan $X$ with $N$ points and fixed model parameter $\theta$, we calculate the difference between the two most likely class labels for all points and produce the score for a point cloud scan ($S_{MAR}$) by averaging the value of all points in a scan. After that, we select a portion of point cloud scans with the largest score in the unlabeled dataset for label acquisition.

\begin{equation}
    S_{MAR} = \frac{1}{N} \sum_{n=1}^N P(\hat{y_n^1} | X;\theta) - P(\hat{y_n^2} | X;\theta),
    \label{eq:mar}
\end{equation}
where $\hat{y_n^1}$ is the first most probable label class and $\hat{y_n^2}$ is the second most probable label class.

\paragraph{Least confidence sampling (CONF)} Many previous active learning methods query the sample whose prediction has the least confidence \cite{ijcnn, wang2016cost}. As can be observed in Eq. \ref{eq:conf}, given a point cloud scan $X$ with $N$ points and fixed model parameter $\theta$, we calculate the confidence of predicted class label ($\hat{y_n^1}$) for all points and produce the score for a point cloud scan ($S_{CONF}$) by averaging the value of all points in a scan. After that, we select a portion of point cloud scans with the least confidence score in the unlabeled dataset for label acquisition.

\begin{equation}
    S_{CONF} = \frac{1}{N} \sum_{n=1}^N P(\hat{y_n^1} | X;\theta)
    \label{eq:conf}
\end{equation}

\paragraph{Softmax entropy (ENT)} Entropy is an indicator to measure the information of a probability distribution in the information theory \cite{shannon1948mathematical}. Some previous active learning approaches query samples with the highest entropy value in the predicted probability \cite{ijcnn}. As shown in Eq. \ref{eq:ent}, given a point cloud scan $X$ with $N$ points and fixed model parameter $\theta$, we calculate the softmax entropy value for all points and produce the score for a point cloud scan ($S_{ENT}$) by averaging the value of all points in a scan. After that, we select a portion of point cloud scans with the largest entropy in the unlabeled dataset for label acquisition.

\begin{equation}
    S_{ENT} = - \frac{1}{N} \sum_{n=1}^N\sum_{i=1}^{c} P(y_n^i | X;\theta) \log P(y_n^i | X;\theta),
    \label{eq:ent}
\end{equation}
where $c$ represents the total number of labels, and $P(y_n^i | X;\theta)$ represents the probability that the model predicts point $n$ as class $i$.

\paragraph{Core-Set (CSET)} Sener~\etal \cite{sener2018coreset} proposed a purely diversity-based deep active selection strategy named Core-Set. The strategy aims to select a small subset so that a model trained on the selected subset has a similar performance to that trained on the whole dataset. The method first extracts the feature of each sample. Then, it selects a small number of samples from the unlabeled dataset that is the furthest away from the labeled dataset in the feature space for label acquisition. In the implementation, we choose the middle layer of the encoder-decoder network as the feature.

\paragraph{Segment entropy (SEGENT)} Lin~\etal \cite {lin2020efficient} proposed \textit {segment entropy} to measure the point cloud information in the deep active learning pipeline. This method assumes that each geometrically related area should share similar semantic annotations. Therefore, it calculates the entropy of the distribution of predicted labels in a small area to estimate model uncertainty. 

\paragraph{MC-Dropout (MCDR)} \cite{gal2016dropout, gal2017deep} combined Bayesian active learning with deep learning, which estimated model uncertainty by Monte Carlo Dropout. In the implementation, we set the dropout rate to 0.3 and perform 10 dropout predictions. Note that since there is no dropout layer in MinkowskiNet \cite{choy2019minkowski}, we did not compare with this baseline when using MinkowskiNet.

\section{Experimental Result}
\label{sec:exp}

\begin{table*}
    \centering
    \begin{tabular}{c|cccccccc}
         \% Labeled Data & RAND & MAR & CONF & ENT & CSET & SEGENT & MCDR & ReDAL (Ours) \\
         \hline \hline
         init. & 27.05 & 28.29 & 28.60 & 27.92 & 28.89 & 29.16 & 28.33 & 27.86\\
         \hline
         5 & 31.39 & 30.07 & 32.14 & 31.02 & 33.24 & 34.55 & 29.30 & \textbf{41.27}\\
         7 & 35.37 & 31.34 & 33.76 & 35.10 & 36.59 & 40.97 & 33.68 & \textbf{47.68}\\
         9 & 40.51 & 33.30 & 38.57 & 40.90 & 37.02 & 42.30 & 40.00 & \textbf{52.34}\\
         11 & 44.50 & 39.75 & 40.60 & 41.51 & 41.42 & 43.07 & 41.65 & \textbf{54.28}\\
         13 & 46.28 & 40.41 & 42.43 & 43.42 & 41.34 & 44.48 & 44.04 & \textbf{57.01}\\
         15 & 49.02 & 40.45 & 44.44 & 45.06 & 41.40 & 45.04 & 45.06 & \textbf{57.97}\\
    \end{tabular}
    \caption{Results of IoU performance (\%) on S3DIS~\cite{armeni2016s3dis} with SPVCNN \cite{tang2020spvconv}.}
    \label{tab:spvcnn_s3dis}
\end{table*}

\begin{table*}
    \centering
    \begin{tabular}{c|ccccccc}
         \% Labeled Data & RAND & MAR & CONF & ENT & CSET & SEGENT  & ReDAL (Ours) \\
         \hline \hline
         init. & 26.59 & 25.20 & 25.52 & 26.60 & 25.60 & 26.30 &  25.63\\
         \hline
         5 & 30.22 & 25.87 & 27.81 & 27.60 & 35.58 & 26.66 &  \textbf{39.45}\\
         7 & 34.76 & 32.40 & 30.25 & 28.91 & 38.88 & 30.45 &  \textbf{44.29}\\
         9 & 38.79 & 36.20 & 32.23 & 35.40 & 40.41 & 39.72 &  \textbf{50.50}\\
         11 & 43.80 & 41.31 & 38.39 & 37.10 & 41.28 & 41.95 &  \textbf{55.11}\\
         13 & 46.13 & 42.28 & 42.10 & 37.42 & 43.63 & 44.66 &  \textbf{56.14}\\
         15 & 48.57 & 43.15 & 42.18 & 40.37 & 47.26 & 45.79 &  \textbf{57.26}\\
    \end{tabular}
    \caption{Results of IoU performance (\%) on S3DIS~\cite{armeni2016s3dis} with MinkowskiNet \cite{choy2019minkowski}.}
    \label{tab:minkunet_s3dis}
\end{table*}

\begin{table*}
    \centering
    \begin{tabular}{c|cccccccc}
         \% Labeled Data & RAND & MAR & CONF & ENT & CSET & SEGENT & MCDR & ReDAL (Ours) \\
         \hline \hline
         init. & 41.84 & 42.39 & 42.98 & 41.90 & 42.19 & 43.18 & 42.92 & 41.87\\
         \hline
         2 & 45.41 & 46.84 & 46.31 & 45.57 & 46.98 & 47.89 & 47.57 & \textbf{51.70}\\
         3 & 52.19 & 49.55 & 50.15 & 51.42 & 52.93 & 52.60 & 50.08 & \textbf{55.83}\\
         4 & 54.76 & 51.66 & 54.46 & 51.85 & 54.57 & 53.60 & 53.56 & \textbf{56.86}\\
         5 & 56.89 & 53.21 & 55.41 & 56.45 & 56.45 & 54.00 & 54.40 & \textbf{58.18}\\
    \end{tabular}
    \caption{Results of IoU performance (\%) on SemanticKITTI \cite{behley2019semantickitti} with SPVCNN \cite{tang2020spvconv}.}
    \label{tab:spvcnn_semantickitti}
\end{table*}

\begin{table*}
    \centering
    \begin{tabular}{c|ccccccc}
         \% Labeled Data & RAND & MAR & CONF & ENT & CSET & SEGENT  & ReDAL (Ours) \\
         \hline \hline
         init. & 37.74 & 38.20 & 37.32 & 37.33 & 36.86 & 37.75 &  37.48 \\
         \hline
         2 & 42.74 & 42.73 & 42.01 & 42.16 & 41.25 & 42.62 &  \textbf{48.88}\\
         3 & 48.82 & 45.07 & 47.37 & 45.77 & 45.15 & 49.51 &  \textbf{55.30}\\
         4 & 52.51 & 47.84 & 49.54 & 49.46 & 49.93 & 51.87 &  \textbf{58.35}\\
         5 & 54.67 & 51.27 & 53.49 & 52.34 & 51.89 & 53.12 &  \textbf{59.76}\\
    \end{tabular}
    \caption{Results of IoU performance (\%) on SemanticKITTI \cite{behley2019semantickitti} with MinkowskiNet \cite{choy2019minkowski}.}
    \label{tab:minkunet_semantickitti}
\end{table*}

Due to space limitations, we show the original experimental results here, which are shown in the line charts of the main paper. Table \ref{tab:spvcnn_s3dis}, \ref{tab:minkunet_s3dis}, \ref{tab:spvcnn_semantickitti}, \ref{tab:minkunet_semantickitti} shows the original data of Figure 5 in the main paper. Table  \ref{tab:perclass_comparison}, \ref{tab:minkunet_distribution} present the original data of Table 1, 2 in the main paper.

\clearpage

\begin{table*}
    \setlength\tabcolsep{3pt}
    \centering
    \begin{tabular}{c|c|cccccccccccccccccccc}
         \parbox[t]{2mm} &  {\multirow{4}{*}{\rotatebox[origin=c]{90}{\footnotesize{mIoU}}}} & {\multirow{4}{*}{\rotatebox[origin=c]{90}{\footnotesize{car}}}} & {\multirow{4}{*}{\rotatebox[origin=c]{90}{\footnotesize{bicycle}}}} & {\multirow{4}{*}{\rotatebox[origin=c]{90}{\footnotesize{motorcycle}}}}& {\multirow{4}{*}{\rotatebox[origin=c]{90}{\footnotesize{truck}}}} & {\multirow{4}{*}{\rotatebox[origin=c]{90}{\footnotesize{other-vehicle}}}} & {\multirow{4}{*}{\rotatebox[origin=c]{90}{\footnotesize{person}}}} & {\multirow{4}{*}{\rotatebox[origin=c]{90}{\footnotesize{bicyclist}}}} & {\multirow{4}{*}{\rotatebox[origin=c]{90}{\footnotesize{motorcyclist}}}} & {\multirow{4}{*}{\rotatebox[origin=c]{90}{\footnotesize{road}}}} & {\multirow{4}{*}{\rotatebox[origin=c]{90}{\footnotesize{parking}}}} & {\multirow{4}{*}{\rotatebox[origin=c]{90}{\footnotesize{sidewalk}}}} & {\multirow{4}{*}{\rotatebox[origin=c]{90}{\footnotesize{other-ground}}}} & {\multirow{4}{*}{\rotatebox[origin=c]{90}{\footnotesize{building}}}} & {\multirow{4}{*}{\rotatebox[origin=c]{90}{\footnotesize{fence}}}} & {\multirow{4}{*}{\rotatebox[origin=c]{90}{\footnotesize{vegetation}}}} & {\multirow{4}{*}{\rotatebox[origin=c]{90}{\footnotesize{trunk}}}} & {\multirow{4}{*}{\rotatebox[origin=c]{90}{\footnotesize{terrain}}}} & {\multirow{4}{*}{\rotatebox[origin=c]{90}{\footnotesize{pole}}}} & {\multirow{4}{*}{\rotatebox[origin=c]{90}{\footnotesize{traffic-sign}}}}\\
         
          & &&&&\\
        \footnotesize{method} & &&&&\\
        &  &&&&\\
        \hline \hline
        \footnotesize{Full} & \footnotesize{61.4} & \footnotesize{95.9} & \footnotesize{20.4} & \footnotesize{63.9} & \footnotesize{70.3} & \footnotesize{45.5} & \footnotesize{65.0} & \footnotesize{78.5} & \footnotesize {0.4} & \footnotesize {93.5} & \footnotesize{50.6} &  \footnotesize{82.0} & \footnotesize{0.2} & \footnotesize{91.2} & \footnotesize{63.8} & \footnotesize{87.2} & \footnotesize{68.5} & \footnotesize{74.3} & \footnotesize{64.4} & \footnotesize{50.1} \\
        \hline
        \footnotesize{RAND} & \footnotesize{54.7} & \footnotesize{94.7} & \footnotesize{9.5} & \footnotesize{45.0} & \footnotesize{\textbf{66.8}} & \footnotesize{38.6} & \footnotesize{52.0} & \footnotesize{47.8} & \footnotesize {0.0} & \footnotesize {90.2} & \footnotesize{38.5} &  \footnotesize{76.1} & \footnotesize{\textbf{1.8}} & \footnotesize{88.3} & \footnotesize{\textbf{55.5}} & \footnotesize{\textbf{87.9}} & \footnotesize{\textbf{64.0}} & \footnotesize{\textbf{76.5}} & \footnotesize{60.2} & \footnotesize{45.6}\\

        \footnotesize{ReDAL} & \footnotesize{\textbf{59.8}} & \footnotesize{\textbf{95.4}} & \footnotesize{\textbf{29.6}} & \footnotesize{\textbf{58.6}} & \footnotesize{63.4} & \footnotesize{\textbf{49.8}} & \footnotesize{\textbf{63.4}} & \footnotesize{\textbf{84.1}} & \footnotesize {\textbf{0.5}} & \footnotesize {\textbf{91.5}} & \footnotesize{\textbf{39.3}} &  \footnotesize{\textbf{78.4}} & \footnotesize{1.2} & \footnotesize{\textbf{89.3}} & \footnotesize{54.4} & \footnotesize{87.4} & \footnotesize{62.0} & \footnotesize{74.1} & \footnotesize{\textbf{63.5}} & \footnotesize{\textbf{49.7}}\\
    \end{tabular}
    \caption{\textbf{Results of IoU performance (\%) with only $5\%$ labeled points.} The table shows that our ReDAL achieve better results on most classes compared with baseline random selection. For some classes of small items and objects with complex boundaries, our ReDAL greatly surpass the random selection baseline and even outperform fully supervised result, such as bicycle and bicyclist.}
    \label{tab:perclass_comparison_supp}
\end{table*}

\begin{table*}
    \setlength\tabcolsep{3pt}
    \centering
    \begin{tabular}{c|c|cccccccccccccccccccc}
         \parbox[t]{2mm} &  {\multirow{4}{*}{\rotatebox[origin=c]{90}{\footnotesize{total}}}} & {\multirow{4}{*}{\rotatebox[origin=c]{90}{\footnotesize{car}}}} & {\multirow{4}{*}{\rotatebox[origin=c]{90}{\footnotesize{bicycle}}}} & {\multirow{4}{*}{\rotatebox[origin=c]{90}{\footnotesize{motorcycle}}}}& {\multirow{4}{*}{\rotatebox[origin=c]{90}{\footnotesize{truck}}}} & {\multirow{4}{*}{\rotatebox[origin=c]{90}{\footnotesize{other-vehicle}}}} & {\multirow{4}{*}{\rotatebox[origin=c]{90}{\footnotesize{person}}}} & {\multirow{4}{*}{\rotatebox[origin=c]{90}{\footnotesize{bicyclist}}}} & {\multirow{4}{*}{\rotatebox[origin=c]{90}{\footnotesize{motorcyclist}}}} & {\multirow{4}{*}{\rotatebox[origin=c]{90}{\footnotesize{road}}}} & {\multirow{4}{*}{\rotatebox[origin=c]{90}{\footnotesize{parking}}}} & {\multirow{4}{*}{\rotatebox[origin=c]{90}{\footnotesize{sidewalk}}}} & {\multirow{4}{*}{\rotatebox[origin=c]{90}{\footnotesize{other-ground}}}} & {\multirow{4}{*}{\rotatebox[origin=c]{90}{\footnotesize{building}}}} & {\multirow{4}{*}{\rotatebox[origin=c]{90}{\footnotesize{fence}}}} & {\multirow{4}{*}{\rotatebox[origin=c]{90}{\footnotesize{vegetation}}}} & {\multirow{4}{*}{\rotatebox[origin=c]{90}{\footnotesize{trunk}}}} & {\multirow{4}{*}{\rotatebox[origin=c]{90}{\footnotesize{terrain}}}} & {\multirow{4}{*}{\rotatebox[origin=c]{90}{\footnotesize{pole}}}} & {\multirow{4}{*}{\rotatebox[origin=c]{90}{\footnotesize{traffic-sign}}}}\\
         
          & &&&&\\
        \footnotesize{method} & &&&&\\
        &  &&&&\\
        \hline \hline
        \footnotesize{Full} & \footnotesize{$10^3$} & \footnotesize{43.68} & \footnotesize{0.17} & \footnotesize{0.41} & \footnotesize{2.02} & \footnotesize{2.40} & \footnotesize{0.36} & \footnotesize{0.13} & \footnotesize {0.04} & \footnotesize {205.22} & \footnotesize{15.19} &  \footnotesize{148.59} & \footnotesize{4.03} & \footnotesize{137.00} & \footnotesize{74.69} & \footnotesize{275.57} & \footnotesize{6.23} & \footnotesize{80.67} & \footnotesize{2.95} & \footnotesize{0.63} \\
        \hline
        \footnotesize{RAND} & \footnotesize{$10^3$} & \footnotesize{43.89} & \footnotesize{0.14} & \footnotesize{0.34} & \footnotesize{3.51} & \footnotesize{2.12} & \footnotesize{0.42} & \footnotesize{0.11} & \footnotesize {0.05} & \footnotesize {206.86} & \footnotesize{14.07} &  \footnotesize{147.32} & \footnotesize{4.02} & \footnotesize{137.63} & \footnotesize{74.47} & \footnotesize{274.47} & \footnotesize{6.21} & \footnotesize{80.54} & \footnotesize{3.02} & \footnotesize{0.73}\\
        
        \footnotesize{ReDAL} & \footnotesize{$10^3$} & \footnotesize{33.71} & \footnotesize{0.25} & \footnotesize{0.51} & \footnotesize{8.01} & \footnotesize{11.36} & \footnotesize{1.27} & \footnotesize{0.21} & \footnotesize {0.07} & \footnotesize {168.16} & \footnotesize{20.15} &  \footnotesize{145.77} & \footnotesize{16.92} & \footnotesize{132.22} & \footnotesize{78.68} & \footnotesize{252.65} & \footnotesize{9.25} & \footnotesize{114.45} & \footnotesize{4.48} & \footnotesize{1.87}\\
    \end{tabular}
    \caption{\textbf{Labeled Class Distribution Ratio (\textperthousand).} With limited annotation budgets, our active method ReDAL queries more labels on small objects like person and bicycle but less on large uniform areas like road and vegetation. The selection strategy can mitigate the label imbalance problem and improve the performance on more complicated object scenes without hurting much on large areas as shown in Table~\ref{tab:perclass_comparison_supp}.}
    \label{tab:minkunet_distribution}
\end{table*}

\clearpage

\end{document}